\documentclass[10pt]{article}
\usepackage[preprint]{tmlr}
\usepackage{amsmath,amssymb,amsfonts,bm}

\def\eqref#1{equation~\ref{#1}}

\def\1{\bm{1}}

\DeclareMathAlphabet{\mathsfit}{\encodingdefault}{\sfdefault}{m}{sl}
\SetMathAlphabet{\mathsfit}{bold}{\encodingdefault}{\sfdefault}{bx}{n}

\usepackage{hyperref}
\usepackage{url}
\usepackage{graphicx}
\usepackage{amsmath,amssymb}
\usepackage{booktabs}
\usepackage{subcaption}
\usepackage{algorithm}
\usepackage{algpseudocode}
\usepackage{rotating}
\usepackage{multirow}

\title{Shortcut Trajectory Planning for Efficient Offline Reinforcement Learning}

\author{\name Guanquan Wang \email guanquan-wang@g.ecc.u-tokyo.ac.jp\\
      \addr Department of Information and Communication Engineering\\
     The University of Tokyo
      \AND
      \name  Yoshimasa Tsuruoka \email yoshimasa-tsuruoka@g.ecc.u-tokyo.ac.jp\\
      \addr Department of Information and Communication Engineering \\
      The University of Tokyo}

\begin{document}

\maketitle

\begin{abstract}
Diffusion-based trajectory planners have shown strong performance in offline reinforcement learning, but their iterative denoising process often incurs high inference cost. Consistency-based planners reduce the number of sampling steps, yet they typically rely on a two-stage teacher--student distillation pipeline that increases training cost and may introduce instability. We propose Shortcut Trajectory Planning (STP), an offline model-based reinforcement learning framework that incorporates shortcut models as efficient trajectory generators. STP trains a conditional shortcut trajectory model in a single stage, supports adjustable one-step and few-step inference through step-size conditioning, and selects candidate plans using a critic augmented with feasibility-aware correction. Across standard D4RL benchmarks, including locomotion, navigation, manipulation, and dexterous control tasks, STP achieves strong performance while simplifying the training pipeline for fast generative planning.
\end{abstract}

\section{Introduction}
\label{sec:stp_intro}

Diffusion-based trajectory planners have shown strong performance in offline reinforcement learning by modeling behavior trajectories as generative distributions and using planning-time sampling for decision making \citep{janner2022planning,ajay2022conditional}. 
However, their iterative denoising process often requires many sampling steps, which makes them computationally expensive for real-time or high-frequency control. 
Recent consistency-based planners, including Consistency Planning (CP) \citep{wang2024planning} and Consistency Trajectory Planning (CTP) \citep{wang2026consistency}, address this limitation by replacing multi-step diffusion sampling with consistency-based generation \citep{song2023consistency,ding2023consistency,kim2024consistency}. 
These methods substantially reduce inference cost while maintaining competitive planning performance.

Despite their fast inference, CP and CTP still rely on a two-stage training pipeline. 
They first train an Elucidated Diffusion Model (EDM) \citep{karras2022elucidating} as a teacher and then distill it into a consistency model \citep{song2023consistency,luo2023latent}. 
This teacher-student procedure increases the total training cost and introduces an additional source of instability: the final planner depends on both a successfully trained teacher and a successfully distilled student. 
Such instability is particularly undesirable in offline RL, where learning is already challenging due to multimodal trajectory distributions and distributional shift \citep{fujimoto2019off,kumar2020conservative,yu2020mopo,kidambi2020morel}.

In this paper, we propose \emph{Shortcut Trajectory Planning} (STP), a shortcut-model-based trajectory planner for offline model-based RL. 
Shortcut models \citep{frans2025one} learn step-size-conditioned updates and can therefore support one-step, few-step, and multi-step generation within a single network. 
Unlike consistency-based planners that require teacher-student distillation, shortcut models can be trained directly in a single stage, simplifying the learning pipeline while preserving the ability to perform fast sampling. 
This makes shortcut models a natural fit for trajectory planning, where different tasks may require different trade-offs between planning speed and trajectory quality.

We evaluate STP on standard offline RL benchmarks \citep{fu2020d4rl} and compare it with diffusion-based, consistency-based, and CTM-based planners. 
Our experiments show that STP achieves competitive planning performance while reducing both inference cost and training complexity. 
These results suggest that shortcut models provide a practical alternative to distillation-based generative planners for efficient offline reinforcement learning.

\section{Related Work}
\label{sec:stp_related_work}

\textbf{Shortcut Models.}
Shortcut models were recently introduced as an efficient generative modeling framework for accelerating diffusion sampling \citep{frans2025one}. 
Instead of learning only infinitesimal denoising dynamics, shortcut models condition the network on the desired step size, enabling one-step, few-step, and multi-step generation within a single model. 
Compared with standard diffusion models \citep{ho2020denoising,song2020score,karras2022elucidating}, this formulation reduces the number of required sampling steps. 
Compared with consistency models \citep{song2023consistency} and consistency trajectory models \citep{kim2023consistency}, it avoids a multi-stage teacher-student distillation pipeline and can be trained in a single stage. 
This combination of training simplicity and adjustable inference cost makes shortcut models attractive for decision-making problems where computation and control precision must be balanced.

\textbf{Shortcut Models in Reinforcement Learning.}
The use of shortcut models in reinforcement learning remains relatively unexplored. 
The closest related work is Scalable Offline Reinforcement Learning (SORL), which applies shortcut models as generative policy representations in offline RL and uses a learned Q-function as an inference-time verifier \citep{espinosa2026scaling}. 
Dreamer 4 also adopts a shortcut forcing objective to train an efficient world model for imagination-based reinforcement learning \citep{hafner2025dreamer4}. 
These works show that shortcut-style objectives can support efficient decision making, but they use shortcut models either as policies or world models.

In contrast, this paper studies shortcut models for offline model-based reinforcement learning with trajectory planning. 
Rather than parameterizing a policy or predicting future latent states, we use shortcut models as trajectory generators inside a planning framework. 
This setting is closely related to Consistency Planning (CP) \citep{wang2024planning} and Consistency Trajectory Planning (CTP) \citep{wang2026consistency}, but replaces their distillation-based consistency pipeline with a single-stage shortcut-based planner. 
The goal is to retain efficient inference while simplifying training and reducing the instability introduced by teacher-student distillation.

\section{Background on Shortcut Models}
\label{sec:background_shortcut}

Shortcut models \citep{frans2025one} are efficient generative models built on flow matching \citep{lipman2023flow,liu2023flow}. 
They retain the continuous transport view of flow matching, but additionally condition the model on the desired step size, enabling one-step, few-step, and multi-step generation within a single network. 
This section briefly introduces the flow-matching notation used in this paper and then summarizes the shortcut formulation.

\textbf{Flow Matching.}
Flow matching learns a velocity field that transports samples from a simple noise distribution to the data distribution. 
Let $\mathbf{x}_0 \sim \mathcal{N}(\mathbf{0}, \mathbf{I})$ be a noise sample and $\mathbf{x}_1 \sim \mathcal{D}$ be a data sample. 
For $t \in [0,1]$, define the linear interpolation
\begin{equation}
    \mathbf{x}_t = (1-t)\mathbf{x}_0 + t\mathbf{x}_1,
    \qquad
    \mathbf{v}_t = \mathbf{x}_1 - \mathbf{x}_0 .
\end{equation}
A flow-matching model $f_\theta(\mathbf{x}_t,t)$ is trained to predict $\mathbf{v}_t$:
\begin{equation}
    \label{eq:flow_matching_loss}
    \mathcal{L}_{\mathrm{FM}}(\theta)
    =
    \mathbb{E}_{\mathbf{x}_0,\mathbf{x}_1,t}
    \left[
        \left\| f_\theta(\mathbf{x}_t,t) - (\mathbf{x}_1 - \mathbf{x}_0) \right\|_2^2
    \right].
\end{equation}
At inference time, samples are generated by numerical integration,
\begin{equation}
    \label{eq:flow_sampling}
    \mathbf{x}_{t+d} = \mathbf{x}_t + d f_\theta(\mathbf{x}_t,t),
\end{equation}
where $d=1/K$ for $K$ sampling steps. 
Although effective, standard flow matching may require many small steps; using large steps can introduce substantial discretization error.

\textbf{Shortcut Models.}
Shortcut models address this issue by learning finite-step updates directly. 
Instead of predicting only an infinitesimal velocity, a shortcut model conditions on the current sample, time, and requested step size:
\begin{equation}
    S_\theta : (\mathbf{x}_t,t,d) \mapsto \Delta \mathbf{x},
    \qquad
    \mathbf{x}'_{t+d} = \mathbf{x}_t + d\,S_\theta(\mathbf{x}_t,t,d).
    \label{eq:shortcut_update}
\end{equation}
When $d$ is small, this recovers the usual flow-matching behavior; when $d$ is large, the model learns to approximate a larger jump along the generative trajectory. 
The central training signal is a recursive self-consistency constraint: a shortcut of size $2d$ should match the composition of two shortcuts of size $d$,
\begin{equation}
    \label{eq:shortcut_self_consistency}
    S(\mathbf{x}_t,t,2d)
    =
    \frac{1}{2}S(\mathbf{x}_t,t,d)
    +
    \frac{1}{2}S(\mathbf{x}'_{t+d},t+d,d).
\end{equation}
Thus, the same model can be sampled with different inference budgets by choosing $d=1/K$:
\begin{equation}
    \label{eq:shortcut_sampling}
    \mathbf{x}_{t+d} = \mathbf{x}_t + d\,S_\theta(\mathbf{x}_t,t,d),
    \qquad t \leftarrow t+d .
\end{equation}

\textbf{Training Objective.}
Shortcut models combine a flow-matching term with a self-consistency term \citep{frans2025one}. 
The former grounds the model at the finest scale, while the latter propagates this behavior to larger step sizes:
\begin{equation}
    \label{eq:shortcut_loss}
    \mathcal{L}_{\mathrm{SC}}(\theta)
    =
    \mathbb{E}_{\mathbf{x}_0,\mathbf{x}_1,t,d}
    \left[
        \left\|S_\theta(\mathbf{x}_t,t,0) - (\mathbf{x}_1 - \mathbf{x}_0)\right\|_2^2
        +
        \left\|S_\theta(\mathbf{x}_t,t,2d) - S_{\mathrm{target}}\right\|_2^2
    \right],
\end{equation}
where
\begin{equation}
    S_{\mathrm{target}}
    =
    \frac{1}{2}S_\theta(\mathbf{x}_t,t,d)
    +
    \frac{1}{2}S_\theta(\mathbf{x}'_{t+d},t+d,d).
\end{equation}
Unlike consistency models and consistency trajectory models, which often rely on teacher-student distillation \citep{song2023consistency,kim2023consistency}, shortcut models can be trained in a single stage with a single network. 
This makes them appealing for trajectory planning: they preserve flexible low-step inference while reducing training complexity, which is important for offline RL settings where diffusion-based planners can be costly at inference time \citep{janner2022planning,ajay2022conditional}.

\paragraph{Implications for planning.}
For the purposes of offline model-based reinforcement learning, 
the shortcut formulation \citep{frans2025one} is particularly appealing because it combines two properties that are desirable in trajectory generation. 
First, it preserves low-latency inference by allowing generation with a small number of function evaluations, 
including the one-step case. 
Second, it permits the sampling budget to be adjusted at test time, 
so that easier control tasks may use fewer steps while more challenging tasks may benefit from additional refinement. 
Compared with teacher-student consistency-based methods \citep{song2023consistency,kim2023consistency}, 
shortcut models also simplify training by learning the generative shortcut within a single end-to-end optimization procedure. 
These properties are particularly relevant to generative planning in offline RL, 
where trajectory-level generative models have been shown to provide expressive planning mechanisms but often suffer from high inference cost \citep{janner2022planning, ajay2022conditional}. 
They therefore motivate the use of shortcut models as trajectory planners in the proposed method of this paper.

\section{Planning with Shortcut Models}
\label{sec:stp}

\begin{figure}[t]
    \centering
    \includegraphics[width=\linewidth,trim=0 5cm 0 4cm,clip=True]{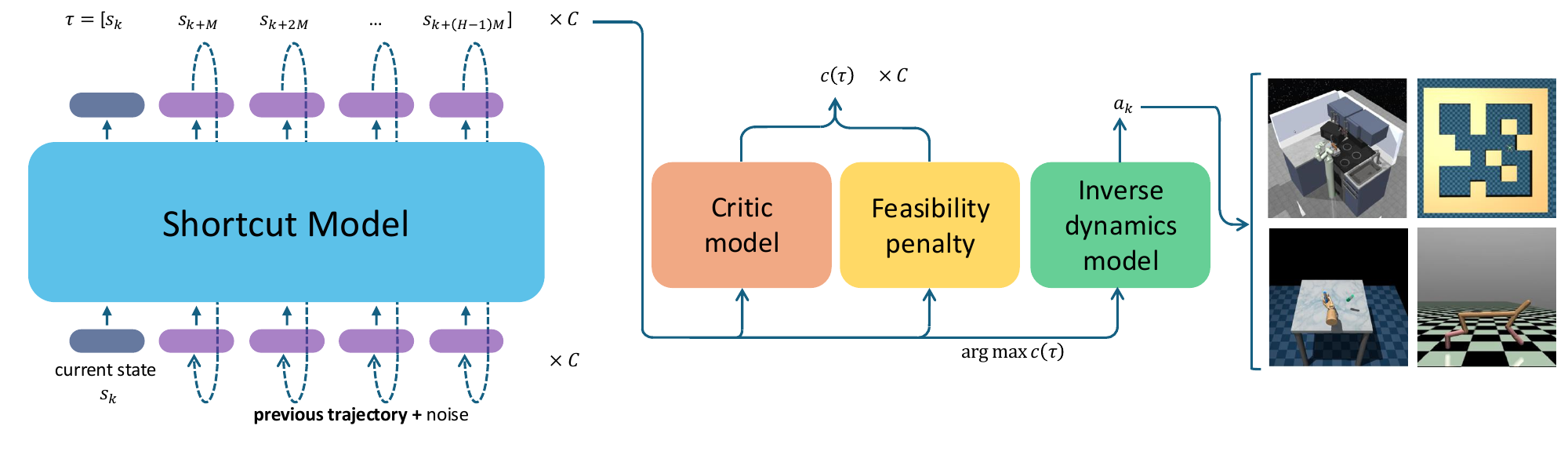}
    \caption{Overview of the proposed Shortcut Trajectory Planning (STP). The shortcut model generates candidate trajectories, which are evaluated by the critic together with a feasibility penalty. The highest-scoring trajectory is converted into the control action using the inverse dynamics model.}
    \label{fig:framework}
\end{figure}

We incorporate shortcut models into offline model-based RL by replacing the distillation-based consistency trajectory generator used in CTP with a single-stage shortcut trajectory model \citep{frans2025one}. 
Figure \ref{fig:framework} provides an overview of the proposed Shortcut Trajectory Planning (STP) framework. The planner generates candidate trajectories using the shortcut model, selects the best trajectory with a learned critic and a feasibility penalty, and recovers the executable action through an inverse dynamics model \citep{janner2022planning,ajay2022conditional,lu2025makes}.

\subsection{Training process}
\label{sec:stp_training_process}

\textbf{Trajectory representation.}
As in CTP \citep{wang2026consistency}, we adopt a trajectory representation with jumping-steps for planning.  
Given a trajectory $\tau$, the planner models a subsequence of future states as
\begin{align}
    \label{eq:shortcut_trajectory_representation}
    \mathbf{x}_{t_i}(\tau) := (s_k, s_{k+M}, \ldots, s_{k+(H-1)M})_{t_i}.
\end{align}
Here, $k$ denotes the timestep index within the trajectory $\tau$, $H$ is the planning horizon, and $M \in \mathbb{N}_{+}$ is the planning stride.  
Thus, the generated subsequence covers $H \times M$ environment steps in total.  
This jump-step planning strategy allows the model to plan over a longer effective horizon while keeping the generated trajectory compact \citep{lu2025makes}.  

Accordingly, $\mathbf{x}_{t_i}(\tau)$ represents a noisy sequence of future states at noise level $t_i$, 
where each element corresponds to one stride-separated state in the planned trajectory.  
During training and inference, the current state $s_k$ is used as the condition, 
and the shortcut model generates the future state sequence $(s_{k+M}, \ldots, s_{k+(H-1)M})$.  
Unlike CTP, where $t_i \in [\epsilon, t_N]$ follows the EDM-style noise parameterization \citep{karras2022elucidating}, 
in this paper we use the flow-matching \citep{lipman2023flow,frans2025one} notation introduced in Section \ref{sec:background_shortcut} and represent the noise level by a continuous variable $t \in [0,1]$.

\textbf{Stride-based inverse dynamics model.}
As in CTP, we employ a stride-based inverse dynamics model to recover actions from the planned state sequence.  
Given the planning stride $M$, the inverse dynamics model $h_{\boldsymbol{\varphi}}$ takes the state pair $(s_k, s_{k+M})$ as input and predicts the corresponding action $a_k$ \citep{agrawal2016learning,pathak2018zero,lu2025makes}.  
Accordingly, during inference, the action is recovered from the selected trajectory via
\[
a_k = h_{\boldsymbol{\varphi}}(s_k, s_{k+M}).
\]
The model is trained on state-action-state tuples sampled from the offline dataset $\mathcal{D}$ by minimizing
\begin{equation}
    \begin{aligned}
        \label{eq:inv_dy_loss_stp}
        \mathcal{L}({\boldsymbol{\varphi}})
        :=
        \mathbb{E}_{(s_k,a_k,s_{k+M}) \sim \mathcal{D}}
        \left[
            \left\|a_k-h_{\boldsymbol{\varphi}}(s_k,s_{k+M})\right\|_2^2
        \right].
    \end{aligned}
\end{equation}

\textbf{Shortcut trajectory model training.}
Unlike the two-stage training procedure adopted in CP and CTP, 
the shortcut trajectory planner is trained in a single stage.  
Specifically, we parameterize the planner as a conditional shortcut model
\begin{equation}
    \label{eq:stp_model}
    S_{\boldsymbol{\theta}}(\mathbf{x}_t, t, d \mid s_k),
\end{equation}
where $\mathbf{x}_t$ denotes the noisy future-state sequence at signal level $t \in [0,1]$, $d \in (0,1]$ is the prescribed step size, and $s_k$ is the current state used as the condition.  
Given $\mathbf{x}_t$, the model predicts a finite-step update along the generative trajectory, and the corresponding next state is defined by
\begin{equation}
    \label{eq:stp_update}
    \mathbf{x}'_{t+d} = \mathbf{x}_t + d\, S_{\boldsymbol{\theta}}(\mathbf{x}_t, t, d \mid s_k).
\end{equation}

Following the flow-matching formulation introduced in Section \ref{sec:background_shortcut}, we construct noisy training samples by linearly interpolating between a Gaussian noise sample $\mathbf{x}_{0}^{\mathrm{noise}} \sim \mathcal{N}(\mathbf{0}, \mathbf{I})$ and a clean trajectory sample $\mathbf{x}_1(\tau)$:
\begin{equation}
    \label{eq:stp_noisy_sample}
    \mathbf{x}_t(\tau) = (1-t)\mathbf{x}_{0}^{\mathrm{noise}} + t\,\mathbf{x}_1(\tau),
\end{equation}
where
\begin{equation}
    \mathbf{x}_1(\tau) := (s_k, s_{k+M}, \ldots, s_{k+(H-1)M}).
\end{equation}
At the infinitesimal-step limit, the shortcut reduces to the local flow direction.  
Accordingly, the model is grounded at the finest scale by the flow-matching objective
\begin{equation}
    \label{eq:stp_fm_loss}
    \mathcal{L}_{\mathrm{FM}}(\boldsymbol{\theta})
    =
    \mathbb{E}_{\tau \sim \mathcal{D},\, \mathbf{x}_{0}^{\mathrm{noise}} \sim \mathcal{N},\, t \sim p(t)}
    \left[
        \left\|
        S_{\boldsymbol{\theta}}(\mathbf{x}_t, t, 0 \mid s_k)
        -
        \bigl(\mathbf{x}_1(\tau)-\mathbf{x}_{0}^{\mathrm{noise}}\bigr)
        \right\|_2^2
    \right].
\end{equation}

To extend the model from infinitesimal updates to finite-step trajectory generation, a recursive self-consistency relation across step sizes is introduced.
In particular, a shortcut associated with step size $2d$ is required to be consistent with the composition of two successive shortcuts associated with step size $d$.  
Let
\begin{equation}
    \label{eq:stp_half_step}
    \mathbf{x}'_{t+d}
    =
    \mathbf{x}_t + d\, S_{\boldsymbol{\theta}}(\mathbf{x}_t, t, d \mid s_k).
\end{equation}
Then the bootstrap target for the larger step size $2d$ is defined as
\begin{equation}
    \label{eq:stp_target}
    S_{\mathrm{target}}
    =
    \frac{1}{2}S_{\boldsymbol{\theta}}(\mathbf{x}_t, t, d \mid s_k)
    +
    \frac{1}{2}S_{\boldsymbol{\theta}}(\mathbf{x}'_{t+d}, t+d, d \mid s_k).
\end{equation}
The corresponding self-consistency loss is
\begin{equation}
    \label{eq:stp_sc_loss}
    \mathcal{L}_{\mathrm{SC}}(\boldsymbol{\theta})
    =
    \mathbb{E}_{\tau \sim \mathcal{D},\, \mathbf{x}_{0}^{\mathrm{noise}} \sim \mathcal{N},\, (t,d) \sim p(t,d)}
    \left[
        \left\|
        S_{\boldsymbol{\theta}}(\mathbf{x}_t, t, 2d \mid s_k)
        -
        \mathrm{sg}\!\left(S_{\mathrm{target}}\right)
        \right\|_2^2
    \right],
\end{equation}
where $\mathrm{sg}(\cdot)$ denotes the stop-gradient operator.

The shortcut trajectory planner is trained by jointly minimizing the flow-matching term and the recursive self-consistency term:
\begin{equation}
    \label{eq:stp_total_loss}
    \mathcal{L}_{\mathrm{STP}}(\boldsymbol{\theta})
    :=
    \mathcal{L}_{\mathrm{FM}}(\boldsymbol{\theta})
    +
    \lambda_{\mathrm{SC}}\,\mathcal{L}_{\mathrm{SC}}(\boldsymbol{\theta}),
\end{equation}
where $\lambda_{\mathrm{SC}}$ controls the contribution of the self-consistency term.

This training objective directly learns a shortcut \citep{frans2025one} mapping for conditional trajectory generation without relying on a separately trained teacher model or an additional distillation stage.  
The flow-matching term anchors the planner to the underlying denoising dynamics at the finest scale, while the recursive self-consistency term propagates this behavior to larger step sizes.  
As a result, the model can generate candidate trajectories under different inference budgets within a single unified framework.

\textbf{Critic model training.}
As in CTP, a critic model is used to rank the candidate trajectories generated by the planner according to their expected returns.  
Let $V_{\boldsymbol{\alpha}}$ denote the critic network with parameters $\boldsymbol{\alpha}$.  
The critic takes the denoised trajectory $\mathbf{x}_{t_0}(\tau)$ as input and predicts the accumulated discounted return $R_k$, defined by
\begin{align}
    \label{eq:culmulated_r_stp}
    R_k = \sum_{h=0}^{t_{\mathrm{end}}} \gamma^h r_{k+h}.
\end{align}
The critic is trained on trajectories sampled from the offline dataset $\mathcal{D}$ by minimizing the mean squared error
\begin{align}
    \label{eq:critic_loss_stp}
    \mathcal{L}_{\mathrm{critic}}(\boldsymbol{\alpha})
    =
    \mathbb{E}_{\tau \sim \mathcal{D}}
    \left[
        \left(
            V_{\boldsymbol{\alpha}}(\mathbf{x}_{t_0}(\tau)) - R_k
        \right)^2
    \right].
\end{align}
At inference time, the critic is used to evaluate the candidate trajectories generated by the shortcut planner, and the trajectory with the highest predicted return is selected for action extraction and execution.

\subsection{Inference process}
\label{sec:stp_inference_process}

\textbf{Sampling with shortcut models.}
At inference time, the shortcut trajectory planner generates future state sequences by iteratively applying the learned shortcut model under a prescribed sampling budget \citep{frans2025one}.  
Let $K$ denote the number of shortcut updates used during planning, and let $d = 1/K$ be the corresponding step size.  
Given the current environmental state $s_k$, sampling aims to produce a trajectory
\[
(s_k, s_{k+M}, \ldots, s_{k+(H-1)M}),
\]
where the first state is fixed to the current observation and the remaining states are generated by the planner.

For the first environmental timestep, sampling starts from a Gaussian noise sample and progressively refines it into a trajectory plan.  
Specifically, an initial noisy trajectory $\mathbf{x}_0^{\mathrm{noise}} \sim \mathcal{N}(\mathbf{0}, \mathbf{I})$ is drawn, and the planner iteratively applies
\begin{align}
    \label{eq:stp_sampling}
    \mathbf{x}_{t+d} \gets \mathbf{x}_t + d\, S_{\boldsymbol{\theta}}(\mathbf{x}_t, t, d \mid s_k),
\end{align}
for $t = 0, d, 2d, \ldots, 1-d$, until the final trajectory sample $\mathbf{x}_1$ is obtained.

For subsequent environmental timesteps, instead of restarting from pure noise, we adopt a warm-start strategy \citep{janner2022planning}.  
More precisely, let $\mathbf{x}_1^{(k-1)}$ denote the denoised trajectory generated at the previous environmental timestep.  
The planner initializes the next trajectory by perturbing $\mathbf{x}_1^{(k-1)}$ with noise, rather than drawing a completely new sample from the Gaussian prior.  
The resulting noisy initialization is then refined by the same shortcut sampling procedure in \eqref{eq:stp_sampling}.  
In this way, the planner reuses information from the previously generated trajectory and updates it into a new trajectory conditioned on the current state.

This warm-start strategy serves two purposes.  
First, it reduces redundant computation by avoiding repeated trajectory generation from pure noise at every environmental step.  
Second, it improves temporal coherence across consecutive planning steps, since the newly generated trajectory remains connected to the previously selected plan.  
As a result, the resulting state sequences and recovered actions are typically smoother over time, which helps reduce unnecessary zigzag behavior during execution.

After trajectory sampling, $C$ candidate trajectories are generated, ranked by the critic model, and the highest-valued trajectory is selected for action extraction using the stride-based inverse dynamics model.

\textbf{Plan selection.}
At each environmental timestep, the shortcut trajectory planner generates $C$ candidate trajectories conditioned on the current state $s_k$.  
Each candidate is evaluated by the critic model $V_{\boldsymbol{\alpha}}$, which estimates the expected return of the generated trajectory \citep{lu2025makes}.  
However, the critic is trained only to predict accumulated discounted returns from trajectories in the offline dataset, and therefore does not explicitly account for whether a generated trajectory is physically feasible in the environment.  
As a result, trajectories that are inconsistent with environment constraints may still receive spuriously high critic scores.

To address this issue, we augment the value-based plan selection rule with a feasibility correction term.  
Let $\mathbf{x}_{t_0}^{(c)}$ denote the $c$-th denoised candidate trajectory, and let $\Psi(\mathbf{x}_{t_0}^{(c)})$ denote a feasibility penalty that measures whether the generated trajectory violates known environment constraints.  
The corrected score used for plan selection is defined as
\begin{equation}
    \label{eq:corrected_plan_score}
    \tilde{V}\!\left(\mathbf{x}_{t_0}^{(c)}\right)
    =
    V_{\boldsymbol{\alpha}}\!\left(\mathbf{x}_{t_0}^{(c)}\right)
    -
    \lambda_{\mathrm{fea}} \, \Psi\!\left(\mathbf{x}_{t_0}^{(c)}\right),
\end{equation}
where $\lambda_{\mathrm{fea}} > 0$ controls the strength of the feasibility correction.  
The best plan is then selected according to
\begin{equation}
    \label{eq:best_plan_selection}
    c^*
    =
    \arg\max_{c \in \{1,\ldots,C\}}
    \tilde{V}\!\left(\mathbf{x}_{t_0}^{(c)}\right).
\end{equation}

\begin{figure}[t]
  \centering

  \includegraphics[
    trim=5cm 5cm 5cm 5cm,
    clip,
    angle=-90,
    width=0.28\textwidth
  ]{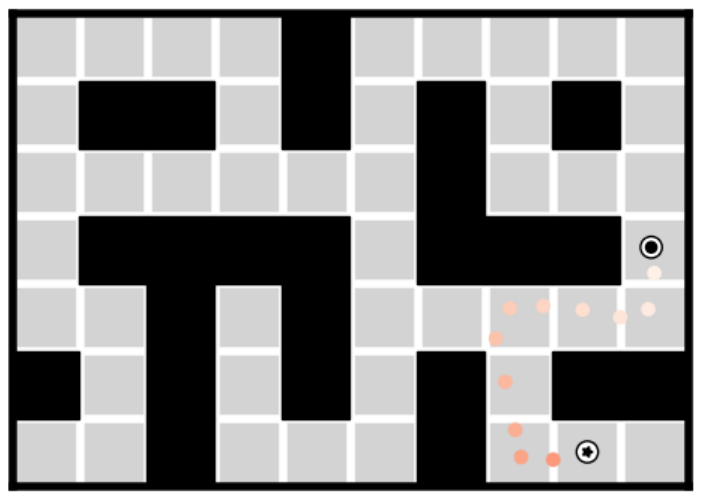}
  \hspace{0.6cm}
  \includegraphics[
    trim=5cm 5cm 5cm 5cm,
    clip,
    angle=-90,
    width=0.28\textwidth
  ]{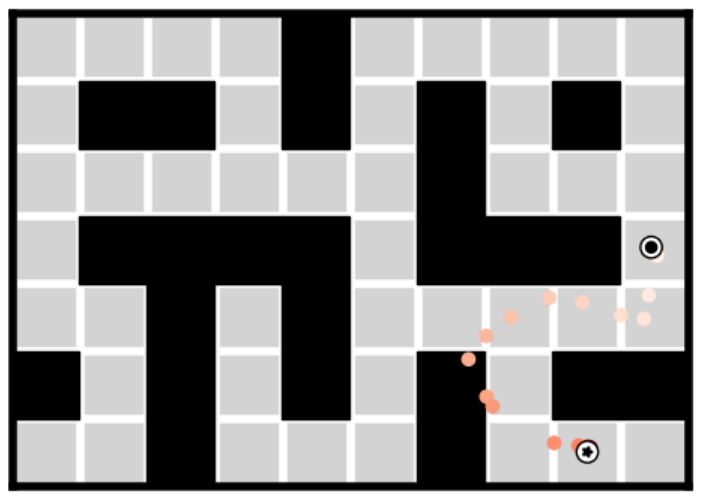}

  \vspace{-0.5em}
  \caption{Examples of generated trajectories in the maze environment. Left: a feasible trajectory whose states remain outside obstacle regions. Right: an infeasible trajectory with intermediate states overlapping obstacle cells. The feasibility-corrected score prevents such unrealizable trajectories from being selected.}
  \label{fig:my_combined_figure}
\end{figure}

For example, as shown in Figure \ref{fig:my_combined_figure}, 
the maze tasks considered in this paper provide a clear illustration of the feasibility correction.  
In this setting, the feasibility penalty is used to detect whether the generated trajectory overlaps with obstacle regions.
Since the agent cannot pass through blocked cells in the true environment, any trajectory that intersects with a wall is infeasible and should not be selected, even if its predicted return is high.  
The feasibility correction therefore prevents the critic from favoring generated trajectories with high predicted cumulative rewards but unrealizable under the environment dynamics.

\textbf{Action extraction.}
After selecting the best trajectory, the action extraction is performed in the same manner as in CTP.  
Specifically, the first stride transition $(s_k, s_{k+M})$ is taken from the selected trajectory, 
and the action is recovered by the stride-based inverse dynamics model \citep{agrawal2016learning,pathak2018zero,lu2025makes}:
\begin{equation}
    \label{eq:stp_action_extraction}
    a_k = h_{\boldsymbol{\varphi}}(s_k, s_{k+M}).
\end{equation}
The resulting action $a_k$ is then executed in the environment, after which the planner replans from the newly observed state. 
The overall procedure is summarized in Algorithm \ref{alg:stp_framework}.

\begin{algorithm}
\small
\caption{Shortcut Trajectory Planning}
\label{alg:stp_framework}
\begin{algorithmic}[1]
\State \textbf{Input}: Planning horizon $H$, dataset $\mathcal{D}$, discount factor $\gamma$, candidate number $C$, planning stride $M$
\State \textbf{Initialize}: Shortcut planner $S_{\boldsymbol{\theta}}$, inverse dynamics model $h_{\boldsymbol{\varphi}}$, critic $V_{\boldsymbol{\alpha}}$
\State Calculate discounted returns $R_k = \sum_{h=0}^{\text{end}} \gamma^h r_{k+h}$ for every step $k$
\Function{Training}{}
    \State Sample $s_k, s_{k+M}, \dots, s_{k+(H-1)M}$, $a_k, a_{k+M}, \dots, a_{k+(H-1)M}$, $R_k$ from $\mathcal{D}$
    \State Construct clean trajectory $\mathbf{x}_1(\tau) := (s_k, s_{k+M}, \dots, s_{k+(H-1)M})$
    \State Sample Gaussian noise $\mathbf{x}_{0}^{\mathrm{noise}} \sim \mathcal{N}(\mathbf{0}, I)$
    \State Sample $t \in [0,1]$ and step size $d$
    \State Construct noisy trajectory $\mathbf{x}_t = (1-t)\mathbf{x}_{0}^{\mathrm{noise}} + t\,\mathbf{x}_1(\tau)$
    \State Train shortcut planner $S_{\boldsymbol{\theta}}$ using Equation \ref{eq:stp_total_loss}
    \State Train inverse dynamics model $h_{\boldsymbol{\varphi}}$ using Equation \ref{eq:inv_dy_loss_stp}
    \State Train critic $V_{\boldsymbol{\alpha}}$ using Equation \ref{eq:critic_loss_stp}
\EndFunction
\Function{Planning}{$s$}
    \If{first environmental timestep}
        \State Generate $C$ candidate plans from Gaussian noise with the first state fixed as $s$
    \Else
        \State Generate $C$ candidate plans using warm-start sampling with the first state fixed as $s$
    \EndIf
    \State Select the best plan using the feasibility-corrected critic score
    \State Generate action with $h_{\boldsymbol{\varphi}}$ using $s$ and the next state in the selected plan
    \State Execute the action in the environment
\EndFunction
\end{algorithmic}
\end{algorithm}

\section{Experiment}
\label{sec:stp_experiment}

We evaluate Shortcut Trajectory Planning (STP) on standard D4RL offline RL benchmarks \citep{fu2020d4rl}, including Gym MuJoCo locomotion tasks \citep{todorov2012mujoco,brockman2016openai}, Maze2D, AntMaze, Kitchen, and Adroit. 
These domains cover continuous control, long-horizon navigation, sparse-reward goal reaching, and high-dimensional manipulation. 
All methods are evaluated in the offline setting without additional environment interaction during training \citep{levine2020offline,fu2020d4rl}.

\subsection{Experimental setting}
\label{sec:stp_exp_setting}

We keep the experimental setting aligned with CTP unless otherwise specified. 
STP is compared with representative actor-critic and generative planning methods, including Diffusion-QL (D-QL) \citep{wang2022diffusion}, Consistency Actor-Critic (C-AC) \citep{ding2023consistency}, Diffuser \citep{janner2022planning}, Decision Diffuser (DD) \citep{ajay2022conditional}, Consistency Planning (CP) \citep{wangplanning}, Consistency Trajectory Planning (CTP), Reward-Aware Consistency Trajectory Distillation (RACTD) \citep{duan2025accelerating}, and Lower Expectile Q-learning (LEQ) \citep{parkmodel}, depending on the benchmark.
We report normalized D4RL returns averaged over 150 independent planning seeds \citep{fu2020d4rl}.

STP uses two shortcut sampling steps for AntMaze and three steps for all other benchmarks. 
For baselines, we follow the sampling budgets in their original implementations: Diffusion-QL uses $N=5$ denoising steps \citep{wang2022diffusion}, Diffuser uses $N=20$ steps, and Decision Diffuser uses $N=40$ steps \citep{janner2022planning,ajay2022conditional}. 
Consistency-based baselines are evaluated with their corresponding reported settings. 
STP generates 30 candidate trajectories per planning step in all experiments except AntMaze-Large, whereas CP and CTP use 50 candidates. 
Warm-start initialization and feasibility-aware plan selection are applied during STP inference, with additional implementation details provided in Appendix \ref{app:stp_details}.

\subsection{Experimental Results}
\label{sec:results}

\begin{table}
    \centering
    \caption{Locomotion results. Prior results are from \citet{ding2023consistency,ajay2022conditional,wang2024planning}.}
    \label{tab:stp_locomotion}
    \begin{tabular}{lcccccc}
        \toprule
        \textbf{Dataset} & \textbf{Diffuser} & \textbf{DD} & \textbf{D-QL} & \textbf{C-AC} &  \textbf{CTP} & \textbf{STP} \\ 
        \midrule
        Halfcheetah-mr & $37.5\pm0.5$ & $39.3\pm4.1$ & $47.8\pm0.3$ & \textbf{58.7}$\pm3.9$ &  $43.4\pm0.4$ & $44.7\pm0.2$ \\ 
        Halfcheetah-m  & $42.8\pm0.3$ & $49.1\pm1.0$ & $51.1\pm0.5$ & \textbf{69.1}$\pm0.7$ & $50.4\pm0.1$ & $51.7\pm0.1$ \\
        \cmidrule(lr){1-7}
        Hopper-mr      & $93.6\pm0.4$ & $100.0\pm0.7$ & \textbf{101.3}$\pm0.6$ & $99.7\pm0.5$ & $90.0\pm1.0$ & $92.6\pm0.1$ \\
        Hopper-m       & $74.3\pm1.4$ & $79.3\pm3.6$ & \textbf{90.5}$\pm4.6$ & $80.7\pm10.5$ & $83.6\pm1.3$ & $\textbf{97.2}\pm0.3$ \\
        \cmidrule(lr){1-7}
        Walker2d-mr    & $70.6\pm1.6$ & $75.0\pm4.3$ & \textbf{95.5}$\pm1.5$ & $79.5\pm3.6$ & $86.9\pm0.3$ & $89.5\pm0.2$ \\
        Walker2d-m     & $79.6\pm0.55$ & $82.5\pm1.4$ & \textbf{87.0}$\pm0.9$ & $83.1\pm0.3$ & $85.7\pm0.2$ & $83.4\pm0.2$ \\
        \midrule
        \textbf{Average} & 63.8 & 70.9 & \textbf{78.9} & 78.5 & 73.3 & 73.9 \\
        \bottomrule
    \end{tabular}
\end{table}


\begin{table}
    \caption{Maze2D results. Prior results are from \citet{janner2022planning,wangplanning,lu2025makes,duan2025accelerating}.}
    \label{tab:stp_maze2d}
    \begin{tabular}{lccccccccc}
        \toprule
        \textbf{Dataset} & \textbf{MPPI} & \textbf{CQL} & \textbf{IQL} & \textbf{Diffuser} & \textbf{CP} & \textbf{RACTD} & \textbf{D-QL} & \textbf{CTP} & \textbf{STP} \\ 
        \midrule
        Maze2D U-Maze & 33.2 & 5.7 & 47.4 & 113.9 & 122.7 & 125.7 & 140.6 & 154.1 & \textbf{159.3} $\pm$ 2.4 \\
        Maze2D Medium & 10.2 & 5.0 & 34.9 & 121.5 & 121.4 & 130.8 & 152.0 & 167.1 & \textbf{176.9} $\pm$ 1.8 \\
        Maze2D Large  & 5.1 & 12.5 & 58.6 & 123.0 & 119.5 & 143.8 & 186.8 & \textbf{216.7} & \textbf{215.1} $\pm$ 2.9 \\
        \midrule
        \textbf{Average} & 16.2 & 7.7 & 47.0 & 119.5 & 121.2 & 133.4 & 159.8 & 179.3 & \textbf{183.8} \\
        \bottomrule
    \end{tabular}
\end{table}
\begin{table}
    \caption{Kitchen results. Prior results are from \citet{ding2023consistency,ajay2022conditional}.}
    \label{tab:stp_kitchen}
    \begin{tabular}{llcccccc}
        \toprule
        \textbf{Dataset} & \textbf{Environment} & \textbf{Diffuser} & \textbf{DD} & \textbf{D-QL} & \textbf{C-AC} & \textbf{CTP} & \textbf{STP} \\ 
        \midrule
        Mixed & Kitchen & 47.5 & $65\pm2.8$ & $62.6\pm5.1$ & $45.8\pm1.5$ & $74.5\pm0.3$ & \textbf{74.7} $\pm$ 0.2 \\
        Partial & Kitchen & 33.8 & $57\pm2.5$ & $60.5\pm6.9$ & $38.2\pm1.8$ & $91.2\pm1.0$ & \textbf{92.5} $\pm$ 1.2 \\
        \midrule
        \textbf{Average} & - & 40.65 & 61.0 & 61.55 & 42.0 & 82.85 & \textbf{83.6} \\
        \bottomrule
    \end{tabular}
\end{table}

As shown in Table \ref{tab:stp_locomotion}, 
STP is evaluated on the medium and medium-replay datasets of three standard D4RL locomotion environments. 
In contrast to the medium-expert setting, 
where near-optimal trajectories are included in the offline dataset, 
the medium and medium-replay datasets mainly contain suboptimal behavior data. 
These settings are therefore more challenging for offline RL and better reveal whether an algorithm can generate improved trajectories from imperfect demonstrations. 
Across these benchmarks, STP achieves competitive results and improves the average score from 73.3 with CTP to 73.9. 
More specifically, STP outperforms CTP on five out of six datasets, 
including both HalfCheetah tasks, both Hopper tasks, and Walker2d-medium-replay. 
The improvement is particularly clear on Hopper-medium, 
where STP achieves the best score among all compared methods. 
These results indicate that the shortcut trajectory formulation can further enhance the planning quality of CTP while retaining the low-step generative inference advantage of consistency-based methods. 
Although Diffusion-QL and Consistency-AC still achieve higher scores on some tasks due to their explicit value-based policy optimization, 
STP remains competitive among trajectory-generation-based planning methods under challenging suboptimal-data settings.

Table \ref{tab:stp_maze2d} shows the performance of STP on the Maze2D benchmark.  
Despite the sparse-reward and long-horizon nature of this domain, STP achieves the best average performance among all compared methods, with an average normalized score of $183.8$.  
It outperforms CTP on Maze2D-Umaze and Maze2D-Medium, and achieves comparable performance on Maze2D-Large.  
These results indicate that the proposed shortcut-based trajectory planner is effective for long-horizon planning and further improves over CTP on average.  
This improvement is also consistent with the intended role of the warm-start strategy and the feasibility-aware plan selection used during inference.

Table \ref{tab:stp_kitchen} presents the performance of STP on the Kitchen benchmark, which evaluates multi-stage manipulation with compositional task structure.  
STP achieves the best performance on both \texttt{kitchen-mixed} and \texttt{kitchen-partial}, and further improves over CTP on both datasets.  
This result indicates that the proposed shortcut-based trajectory planner is effective not only for navigation, but also for compositional robotic manipulation tasks.

\begin{table}[t]
\caption{AntMaze results. Prior results are from \citet{wang2022diffusion,parkmodel}.}
    \label{tab:stp_antmaze}
    \begin{center}
        \begin{tabular}{llllll}
            \multicolumn{1}{l}{\bf Dataset} &\multicolumn{1}{l}{\bf Environment} &\multicolumn{1}{l}{\bf LEQ} &\multicolumn{1}{l}{\bf D-QL} &\multicolumn{1}{l}{\bf CTP} &\multicolumn{1}{l}{\bf STP}
            \\ \hline \\
            Diverse& Antmaze-Large  &$60.2\pm18.3$ &$56.6\pm7.6$ &$82.0\pm3.1$ &$\textbf{83.3}\pm3.0$ \\
            Play&Antmaze-Large &$62.0\pm9.9$ &$46.4\pm8.3$& $\textbf{82.0}\pm3.1$ &$81.3\pm3.2$\\
            Diverse&Antmaze-Medium  &$46.2\pm23.2$ &$78.6\pm10.3$&$86.0\pm2.8$ &$\textbf{88.0}\pm2.7$ \\
            Play&Antmaze-Medium  &$76.3\pm17.2$&$76.6\pm10.8$ &$83.3\pm3.0$ &$\textbf{88.7}\pm2.6$ \\
            \\ \hline \\
            \textbf{Average}&-&61.18&64.6&83.33&\textbf{85.33}\\
        \end{tabular}
    \end{center}
\end{table}

Table \ref{tab:stp_antmaze} reports the performance of STP on the AntMaze benchmark, which evaluates goal-conditioned control under sparse rewards, long-horizon planning, and high-dimensional actuation.  
As shown in Table \ref{tab:stp_antmaze}, STP achieves the best average performance among all compared methods.  
Compared with CTP, STP improves the average score from $83.33$ to $85.33$.  
These results indicate that the proposed shortcut-based trajectory planner remains effective in challenging goal-conditioned control tasks that require both long-horizon planning and high-dimensional continuous control.

\begin{table}[t]
\caption{Adroit expert results. Prior results are from \citet{he2024aligniql,lu2025makes}.}
    \label{tab:stp_adroit}
    \begin{center}
        \begin{tabular}{llcccccccc}
            \multicolumn{1}{l}{\bf Environment} &\multicolumn{1}{l}{\bf BC} &\multicolumn{1}{l}{\bf BCQ} &\multicolumn{1}{l}{\bf CQL} &\multicolumn{1}{l}{\bf IQL} &\multicolumn{1}{l}{\bf AlignIQL} &\multicolumn{1}{l}{\bf D-QL}&\multicolumn{1}{l}{\bf CTP} &\multicolumn{1}{l}{\bf STP}
            \\ \hline \\
            Door &34.9 &99.0 &101.5 &103.8 &104.6 &$104.3$&104.1 &$\textbf{105.3}\pm0.4$  \\
            Hammer &\textbf{125.6} &107.2 &86.7 &116.3 &124.7&55.9 &110.5 &$120.9\pm2.2$\\
            Pen &85.1 &114.9 &107.0 &111.7 &116.0 &60.9&104.1 &$\textbf{122.0}\pm4.7$ \\
            Relocate &101.3 &41.6 &95.0 &102.7 &106.0 &108.8&108.8 &$\textbf{108.9}\pm0.6$ \\
            \\ \hline \\
            \textbf{Average}&86.7 &90.7 &97.6 &108.6 &112.8&82.4 &106.9 &\textbf{114.3}\\
        \end{tabular}
    \end{center}
\end{table}

We further evaluate STP on the expert dataset from the Adroit benchmark, which consists of dexterous manipulation tasks requiring precise control in high-dimensional action spaces.  
Quantitative results are reported in Table \ref{tab:stp_adroit}.  
While STP does not outperform all prior methods on every task, it achieves the best average performance overall.  
These results suggest that STP is effective in goal-conditioned manipulation scenarios that require precise spatial reasoning and fine-grained control.

Overall, the experimental results across D4RL locomotion, Maze2D, Kitchen, AntMaze, and Adroit consistently demonstrate the effectiveness of STP over diverse offline RL benchmarks.  
These environments cover substantially different task structures, 
including long-horizon maze navigation, sparse-reward quadrupedal control, 
compositional manipulation, and high-dimensional dexterous manipulation.  
The consistent performance across these heterogeneous domains and dataset types suggests that the effectiveness of STP is not merely due to overfitting to a specific environment, dataset, or task structure.  
Instead, the results indicate that the shortcut-based trajectory generation mechanism provides a generally effective planning component for offline model-based reinforcement learning.

Compared with CTP, the proposed shortcut-based trajectory planner preserves strong planning performance while achieving improved average results on most domains, indicating that replacing the teacher-student consistency trajectory generator with a single-stage shortcut model does not compromise control quality.  
These findings suggest that shortcut-based trajectory generation provides an effective and robust alternative for offline model-based reinforcement learning.  
In the next subsection, we further examine the contribution of individual design choices in STP through ablation studies.

\subsection{Ablation study}
\label{sec:ablation}
This section presents ablation studies to analyze the contributions of two key design choices in STP: 
the warm-start strategy and the feasibility penalty. 
The former is designed to improve planning efficiency and temporal consistency during online execution, 
while the latter aims to suppress infeasible trajectories in long-horizon Maze2D planning. 
For each component, we conduct controlled comparisons by modifying only the target component while keeping the remaining hyperparameters unchanged.

\textbf{Effect of the warm-start strategy.}
We first examine the effect of the warm-start strategy introduced in Section \ref{sec:stp_inference_process}.  
Warm-start is designed to reuse the trajectory generated at the previous environmental time step, 
rather than restarting the planning process from pure noise at every step.  
This design is expected to improve temporal consistency across consecutive planning steps and allow the planner to refine previously generated trajectories instead of regenerating them from scratch.

\begin{table}[t]
\caption{Ablation study on the warm-start strategy in the Maze2D environment. All other hyperparameters are kept unchanged.}
\label{tab:warm_start_maze2d}
\begin{center}
\begin{tabular}{lcc}
\toprule
\textbf{Dataset} & \textbf{Without warm-start} & \textbf{With warm-start} \\
\midrule
Maze2D U-Maze  & $119.4 \pm 4.4$ & $\mathbf{159.3} \pm 2.4$ \\
Maze2D Medium  & $145.7 \pm 3.1$ & $\mathbf{176.9} \pm 1.8$ \\
Maze2D Large   & $187.2 \pm 4.7$ & $\mathbf{215.1} \pm 2.9$ \\
\midrule
Average        & $150.8$         & $\mathbf{183.8}$ \\
\bottomrule
\end{tabular}
\end{center}
\end{table}

\begin{table}[t]
\caption{Ablation study on the warm-start strategy in the AntMaze environment. All other hyperparameters are kept unchanged.}
\label{tab:warm_start_antmaze}
\begin{center}
\begin{tabular}{lcc}
\toprule
\textbf{Dataset} & \textbf{Without warm-start} & \textbf{With warm-start} \\
\midrule
AntMaze Medium-Diverse & $64.7 \pm 3.9$ & $\mathbf{88.0} \pm 2.7$ \\
AntMaze Medium-Play    & $62.0 \pm 4.0$ & $\mathbf{88.7} \pm 2.6$ \\
AntMaze Large-Diverse  & $78.0 \pm 3.4$ & $\mathbf{83.3} \pm 3.0$ \\
AntMaze Large-Play     & $67.3 \pm 3.8$ & $\mathbf{81.3} \pm 3.2$ \\
\midrule
Average                & $68.0$         & $\mathbf{85.3}$ \\
\bottomrule
\end{tabular}
\end{center}
\end{table}

Tables \ref{tab:warm_start_maze2d} and \ref{tab:warm_start_antmaze} report the results on Maze2D and AntMaze, respectively.  
On Maze2D, warm-start improves the average normalized score from $150.8$ to $183.8$.  
On AntMaze, the average score increases from $68.0$ to $85.3$.  
The improvement is consistent across all tasks in both domains, indicating that warm-start initialization plays an important role in improving planning performance.

By reusing previous trajectories, warm-start makes the generated plans and extracted actions smoother over time, which reduces unnecessary oscillatory behavior and improves long-horizon execution.  
The consistent gains observed in both Maze2D and AntMaze suggest that this effect is not limited to relatively simple navigation domains, but also extends to more challenging environments with sparse rewards and high-dimensional actuation.

\textbf{Effect of the feasibility penalty.}
We then evaluate the effect of the feasibility-aware correction introduced in Section \ref{sec:stp_inference_process}. 
The comparison is conducted on the Maze2D benchmark with and without the feasibility penalty, so that the only difference between the compared variants lies in whether infeasible trajectories are explicitly penalized during plan selection.

The results, shown in Table \ref{tab:feasibility_penalty_maze2d}, indicate that the feasibility penalty has limited effect on Maze2D U-Maze and Maze2D Medium, but yields a substantial improvement on Maze2D Large.  
In particular, the score on Maze2D Large increases from $181.9 \pm 4.7$ to $215.1 \pm 2.9$, whereas the gains on U-Maze and Medium are comparatively small.  
These results suggest that the feasibility-aware correction becomes increasingly important as the planning problem becomes more complex.

In relatively simple tasks such as U-Maze and Medium, 
generated candidate trajectories are less likely to violate environment constraints, 
and therefore value-based plan selection alone is often sufficient.  
In contrast, in the more challenging Maze2D Large setting, 
generated trajectories are more likely to contain infeasible segments that overlap with obstacle regions.  
Since the critic is trained to predict expected return rather than physical feasibility, 
such unrealizable trajectories may still receive high scores and be mistakenly selected.  
By explicitly penalizing trajectories that violate environment constraints, 
the feasibility penalty improves the quality of plan selection and leads to a clear performance gain on the more complex task.

\begin{table}[t]
\caption{Ablation study on the feasibility penalty in the Maze2D environment. All other hyperparameters are kept unchanged.}
\label{tab:feasibility_penalty_maze2d}
\begin{center}
\begin{tabular}{lcc}
\toprule
\textbf{Dataset} & \textbf{Without feasibility penalty} & \textbf{With feasibility penalty} \\
\midrule
Maze2D U-Maze  & $158.5 \pm 2.3$ & $\mathbf{159.3 \pm 2.4}$ \\
Maze2D Medium  & $175.4 \pm 1.6$ & $\mathbf{176.9 \pm 1.8}$ \\
Maze2D Large   & $181.9 \pm 4.7$ & $\mathbf{215.1 \pm 2.9}$ \\
\midrule
Average        & $171.9$         & $\mathbf{183.8}$ \\
\bottomrule
\end{tabular}
\end{center}
\end{table}

\begin{figure}[t]
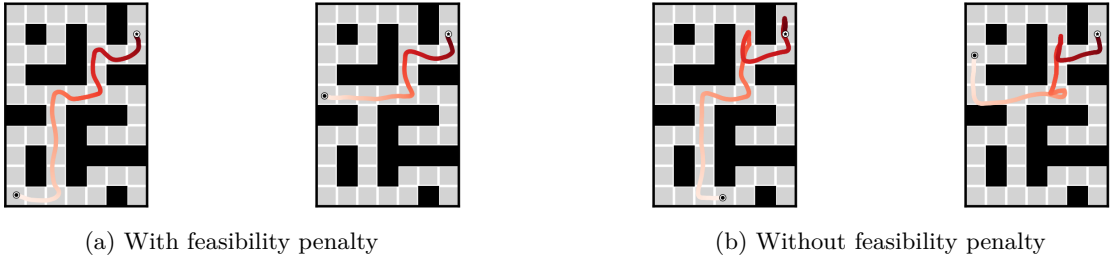

    \centering
    \begin{subfigure}{0.48\textwidth}
        \centering
        \includegraphics[width=0.48\linewidth]{traj_2_w_penalty.pdf}
        \hfill
        \includegraphics[width=0.48\linewidth]{traj_5_w_penalty.pdf}
        \caption{With feasibility penalty}
        \label{fig:maze2d_large_with_penalty}
    \end{subfigure}
    \hfill
    \begin{subfigure}{0.48\textwidth}
        \centering
        \includegraphics[width=0.48\linewidth]{traj_3_wo_penalty.pdf}
        \hfill
        \includegraphics[width=0.48\linewidth]{traj_5_wo_penalty.pdf}
        \caption{Without feasibility penalty}
        \label{fig:maze2d_large_without_penalty}
    \end{subfigure}

    \caption{Qualitative comparison of executed trajectories on Maze2D-large with and without the feasibility penalty. The feasibility penalty encourages smoother routes with fewer obstacle interactions, while value-based selection alone may produce more tortuous paths and wall collisions.}
    \label{fig:maze2d_large_penalty}
\end{figure}

Figure \ref{fig:maze2d_large_penalty} further provides a qualitative comparison of the executed trajectories on Maze2D-large with and without the feasibility penalty. 
As shown in the figure, when the feasibility penalty is applied, 
the agent follows a relatively smooth route from the start position to the goal, 
with fewer unnecessary detours and fewer interactions with obstacle boundaries. 
In contrast, without the feasibility penalty, the agent can still eventually reach the goal, 
but the executed trajectory becomes more tortuous and includes segments that collide with or slide along the maze walls. 
This suggests that value-based plan selection alone may be sufficient for reaching the goal in some cases, 
but it does not necessarily guarantee physically feasible or smooth execution. 
By explicitly accounting for environment constraints, 
the feasibility penalty helps filter out unrealistic candidate trajectories and improves the executability of the selected plans in complex maze layouts.

Overall, the ablation results indicate that the two inference-time designs in STP contribute in complementary ways.  
The warm-start strategy mainly improves temporal consistency across consecutive planning steps, leading to smoother trajectory refinement and more stable long-horizon execution.  
In contrast, the feasibility penalty mainly improves plan selection in more complex environments by suppressing unrealizable trajectories that may otherwise be overestimated by the critic.  
Together, these results support the effectiveness of the proposed inference design in improving the robustness and performance of STP.

\section{Discussion}
\label{sec:stp_discussion}

This paper introduced Shortcut Trajectory Planning (STP), a shortcut-model-based trajectory planner for offline reinforcement learning. 
By replacing the teacher-student distillation pipeline used in consistency-based planners with a single-stage shortcut model, STP simplifies training while preserving efficient few-step trajectory generation and an adjustable inference budget.

Across standard offline RL benchmarks, STP achieves strong performance on sparse-reward navigation, compositional robotic manipulation, high-dimensional goal-conditioned control, and dexterous manipulation. 
The results show that shortcut-based trajectory generation can match or improve over CTP on average, suggesting that removing distillation does not weaken planning capability. 
The ablations further indicate that warm-start sampling and feasibility-aware plan selection are complementary: warm-start improves temporal consistency across planning steps, while feasibility correction suppresses unrealizable trajectories during candidate selection.

Overall, STP shows that shortcut models are a practical alternative to distillation-based consistency models for fast generative planning. 
Its combination of single-stage training, efficient trajectory generation, and robust inference-time selection makes shortcut-based planning a promising direction for offline model-based reinforcement learning.

\bibliography{tmlr}
\bibliographystyle{tmlr}
\clearpage
\appendix
\section{Additional Details on Shortcut Trajectory Planning}
\label{app:stp_details}

This appendix provides additional experimental and implementation details for Shortcut Trajectory Planning (STP), including the sources of baseline results, model architecture, training hyperparameters, planning settings, inference settings, and ablation configurations.

\subsection{Baseline Results and Sources}
\label{app:stp_baseline_sources}

For the benchmark results reported in the main paper, we use previously published scores when the corresponding methods have already been evaluated on the same standardized D4RL tasks.

For the Gym MuJoCo locomotion tasks, the results of Diffuser and Decision Diffuser (DD) are taken from Table 1 of \citep{ajay2022conditional}. 
The results of Diffusion-QL (D-QL) and Consistency-Actor-Critic (C-AC) are taken from Table 2 of \citep{ding2023consistency}. 
The results of Consistency Planning (CP) and Consistency Trajectory Planning (CTP) are taken from their corresponding experimental settings and evaluation protocols.

For the Maze2D tasks, the results of MPPI, CQL, IQL, and Diffuser are taken from Table 1 of \citep{janner2022planning}. 
The results of Diffusion-QL are taken from Table 1 of \citep{lu2025makes}, and the results of RACTD are taken from Table 2 of \citep{duan2025accelerating}. 
The results of CP and CTP are taken from their corresponding experiments under the same benchmark setting.

For the AntMaze, Kitchen, and Adroit tasks, baseline results are reported using the same sources and evaluation protocol as the CTP baseline. 
When a baseline result is reproduced using an existing implementation rather than quoted from a published table, we explicitly follow the implementation and evaluation settings described in the corresponding experiment section.

\subsection{Implementation Details}
\label{app:stp_implementation}

\paragraph{Model architecture.}
We use Diffusion Transformer blocks with the adaLN-Zero architecture \citep{peebles2023scalable} as the network backbone for the shortcut trajectory model. 
The critic and inverse dynamics models follow the same architectural design as the CTP baseline, so that the comparison between CTP and STP focuses on the trajectory generator rather than auxiliary model differences.

\paragraph{Shortcut trajectory model training.}
The shortcut trajectory model is trained in a single stage using the joint objective described in Section \ref{sec:stp_training_process}. 
Unlike CP and CTP, no separately trained diffusion or EDM teacher model is required. 
The training objective consists of a flow-matching term and a recursive self-consistency term across step sizes. 
The model is optimized using Adam.

\paragraph{Critic and inverse dynamics training.}
The critic model and the stride-based inverse dynamics model are trained using the same procedure as the CTP baseline. 
The critic is trained to predict the accumulated discounted return of a denoised candidate trajectory, and the inverse dynamics model is trained to recover the executable action from the first stride transition of the selected trajectory.

\paragraph{Inference settings.}
Unless otherwise specified, STP uses a small number of shortcut sampling steps during inference. 
We use $N=2$ shortcut sampling steps for the AntMaze tasks and $N=3$ steps for all other tasks, including the D4RL locomotion, Maze2D, Kitchen, and Adroit tasks. 
These choices are made to ensure a consistent and fair comparison within each benchmark group. 
In preliminary experiments, we found that some relatively simple environments can already achieve near-saturated performance with fewer shortcut sampling steps; for example, Walker2d and Hopper in the D4RL locomotion benchmark can perform well with $N=2$, and AntMaze-medium can achieve near-saturated scores with $N=1$. 
Nevertheless, we use a unified sampling budget within the corresponding experiment groups to avoid introducing additional task-specific tuning when reporting results in the same table.

\paragraph{Planning horizon and stride.}
The planning horizon $H$ and stride $M$ are set as follows:
\begin{itemize}
    \item \textbf{MuJoCo locomotion tasks:} $H=4$, $M=1$.
    \item \textbf{Kitchen tasks:} $H=32$, $M=4$.
    \item \textbf{Maze2D tasks:} $H=32$, $M=15$.
    \item \textbf{AntMaze tasks:} $H=40$, $M=25$.
    \item \textbf{Adroit tasks:} $H=32$, $M=2$.
\end{itemize}
These settings are kept consistent with the CTP baseline to enable a controlled comparison between CTP and STP. 
In particular, the same horizon and stride are used so that the main difference lies in the generative trajectory model: CTP uses a distilled consistency trajectory model, whereas STP uses a single-stage shortcut trajectory model.

\subsection{Ablation Study Details}
\label{app:stp_ablation_details}

\paragraph{Warm-start strategy.}
For the warm-start ablation, we compare STP with and without reusing the previously generated denoised trajectory as the initialization for the next environmental time step. 
In the variant without warm-start, each planning step starts from an independently sampled Gaussian noise trajectory. 
In the variant with warm-start, the first planning step starts from a pure Gaussian noise trajectory $\mathbf{x}_0 \sim \mathcal{N}(\mathbf{0},\mathbf{I})$. 
For subsequent environmental time steps, the denoised trajectory from the previous step is perturbed by Gaussian noise and reused as the initialization. 
Specifically, we initialize the shortcut sampling process from an intermediate noise level $t=0.3$ by mixing the previous denoised trajectory with Gaussian noise, i.e.,
\[
    \mathbf{x}_{0.3}^{(k)}
    =
    0.7\,\mathbf{x}_{0}^{\mathrm{noise}}
    +
    0.3\,\mathbf{x}_{1}^{(k-1)},
    \qquad
    \mathbf{x}_{0}^{\mathrm{noise}} \sim \mathcal{N}(\mathbf{0},\mathbf{I}),
\]
where $\mathbf{x}_{1}^{(k-1)}$ denotes the denoised trajectory generated at the previous environmental time step. 
Shortcut sampling is then applied from $t=0.3$ to $t=1$ conditioned on the current state. 
All other hyperparameters, including the number of candidate trajectories, sampling steps, planning horizon, stride, critic model, and inverse dynamics model, are kept unchanged.

\paragraph{Feasibility penalty.}
For the feasibility-penalty ablation, we compare value-based plan selection with and without the feasibility correction term introduced in Section \ref{sec:stp_inference_process}. 
In the variant without the feasibility penalty, candidate trajectories are selected only according to the critic score. 
In the variant with the feasibility penalty, the corrected score subtracts a penalty for trajectories that violate known environment constraints. 
We set the feasibility-penalty coefficient to $\lambda_{\mathrm{fea}}=5$ in all experiments using this correction. 
This comparison is mainly conducted on Maze2D, where obstacle regions provide explicit feasibility constraints and infeasible generated trajectories can be directly detected.

\paragraph{Evaluation protocol.}
All ablation experiments use the same evaluation protocol as the main experiments. 
For each task, the reported score is the average normalized return over multiple evaluation episodes. 
Unless otherwise specified, the model checkpoints, critic, inverse dynamics model, planning horizon, stride, and sampling budget are kept fixed across the compared variants.

\end{document}